\documentclass[11pt]{article}
\usepackage[round]{natbib}
\usepackage{latexsym}
\usepackage{amssymb}
\usepackage{url}
\title{\bf Lecture Notes on Fair Division\thanks{Notes prepared for a tutorial at the 11th European Agent Systems Summer School (EASSS-2009), Torino, Italy, 31 August and 1 September 2009. Updated for a tutorial at the COST-ADT Doctoral School on Computational Social Choice, Estoril, Portugal, 9--14 April 2010. Feedback welcome: \texttt{ulle.endriss@uva.nl}.}}
\author{Ulle Endriss}
\date{Institute for Logic, Language and Computation \\ University of Amsterdam \\[12pt] 7 April 2010}

\edef\today{\number\day\space
\ifcase\month\or
January\or February\or March\or April\or May\or June\or July\or
August\or September\or October\or November\or December\fi\ \number\year}

\usepackage{natbib}
\newtheorem{definition}{Definition}[section]

\newtheorem{lemma}{Lemma}[section]
\newtheorem{axiom}{Axiom}

\newtheorem{exer}{Exercise}[section]
\newenvironment{exercise}{\begin{exer}\rm }{\end{exer}}

\newtheorem{examp}{Example}[section]
\newenvironment{example}{\begin{examp}\rm }{\nobreak\hfill~$\Box$\end{examp}}

\newenvironment{cakeproc}[2]%
{\begin{center}\begin{tabular}{|c|}\hline
\textsc{#1} (#2 players) \\ \hline\hline \\[-9pt]
\begin{minipage}{12.8cm}}%
{\end{minipage} \\[12pt] \hline
\end{tabular}\end{center}}

\newcommand{\sw}[1]{\mbox{\rm SW}_{\scriptsize{\rm #1}}}
\newcommand{\OWA}{\mbox{\rm OWA}}
\newcommand{\preceqlex}{\preceq_{\scriptsize{\rm lex}}}
\newcommand{\preclex}{\prec_{\scriptsize{\rm lex}}}
\newcommand{\R}{\mathbb{R}}

\begin{document}
\maketitle

\begin{abstract}\noindent
Fair division is the problem of dividing one or several goods amongst two or more agents in a way that satisfies a suitable fairness criterion.
These Notes provide a succinct introduction to the field. 
We cover three main topics.
First, we need to define what is to be understood by a ``fair'' allocation of goods to individuals. We present an overview of the most important \emph{fairness criteria} (as well as the closely related criteria for economic efficiency) developed in the literature, 
together with a short discussion of their axiomatic foundations.
Second, we give an introduction to \emph{cake-cutting procedures} as an example of methods for fairly dividing a single \emph{divisible} resource amongst a group of individuals.
Third, we discuss the \emph{combinatorial optimisation} problem of fairly allocating a set of \emph{indivisible} goods to a group of agents, covering both centralised algorithms (similar to auctions) and a distributed approach based on negotiation.

While the classical literature on fair division has largely developed within Economics, these Notes are specifically written for readers with a background in Computer Science or similar, and who may be (or may wish to be) engaged in research in Artificial Intelligence, Multiagent Systems, or Computational Social Choice.
References for further reading, as well as a small number of exercises, are included.
\end{abstract}

\clearpage
\tableofcontents
\clearpage

\section{Introduction}

Fair division is the problem of dividing one or several goods amongst two or more agents in a way that satisfies a suitable fairness criterion. That is, fair division may be considered part of the larger research area of \emph{multiagent resource allocation} \citep{mara-survey}. What is special about fair division is the explicit focus on fairness concerns.

These notes give a succinct introduction to the field, focusing on formal and computational aspects that are particularly relevant to research in Computational Social Choice \citep{ChevaleyreEtAlSOFSEM2007} and Multiagent Systems \citep{Wooldridge2009}. We begin by briefly outlining how fair division fits into (and relates to) these two disciplines.

Like voting, the archetypical instance of a social choice problem, fair division amounts to selecting an outcome from a set of possible collective agreements, given the individual preferences of a group of agents. There are however two main differences when compared to voting. The first difference is that, typically, voting theory assumes that agents (voters) have ordinal preferences (that is, they rank the available candidates and can say for any two candidates which one they like more), while in the context of fair division we usually assume that agents have cardinal preferences (that is, each agent has got a utility function mapping possible outcomes to appropriate numerical values). The second difference is that a fair division problem comes with a certain internal ``structure'' that is typically absent from problems in voting:\footnote{This is true for classical voting theory; the situation is somewhat different for voting in combinatorial domains \citep{ChevaleyreEtAlAIMag2008}.} In fair division, the set of possible collective agreements is the set of possible allocations of goods (or parts thereof) to agents; if two distinct allocations assign the same bundle to a particular agent, it is reasonable to assume that that agent will be indifferent between the two allocations in question. In voting theory, on the other hand, collective agreements are just names of candidates (to be elected), and we cannot infer anything regarding the relative preferences of a given voter over two given candidates by just looking at the names of these candidates. Having said this, at a more abstract, level fair division and voting share a number of characteristics, and the computationally oriented study of both problems are part of the field of Computational Social Choice.  

Multiagent resource allocation, and thereby also fair division, are also central topics in Multiagent Systems: First, much work in Multiagent Systems is itself directed towards designing mechanisms for resource allocation. Second, in most other cases resource allocation arises at least as an important side issue (e.g., in collaborative problem solving the agents may have to first agree on a suitable division of the resources available to them, so that each individual agent can carry out the tasks assigned to them successfully). Finally, many of the typical application areas studied in the Multiagent Systems research community are also closely related to multiagent resource allocation (e.g., electronic commerce).

Fairness often plays an important role: a user may only be willing to have their software agent interact with other agents in a system provided by a third party, if the developers of that system can give some fairness guarantees; the developers of an electronic commerce platform may wish to ensure some basic fairness properties for their customers so as to not lose their custom in the future; and so forth.   

But the study of fairness and of fair division long predates modern interest in multiagent systems. Of particular importance are the early contributions by mathematicians of the \emph{Polish School} in the 1940s (Banach, Knaster, Steinhaus). They were the first to recognise the problem of fairly dividing a cake between several agents as a question of considerable mathematical interest. We will recall some of their ideas in these Notes. Cakes are examples for divisible goods: we can cut them in as many pieces as desired and allocate pieces of any size to an agent. When indivisible goods are concerned, each of which needs to be allocated as a whole to a single agent, the nature of the problem changes. It may then be regarded as a combinatorial optimisation problem. To date, most work on resource allocation in multiagent systems has been of this latter type, and we will review some of it here. But before discussing algorithms for fair division, we first need to settle what we actually mean when we say that an allocation is ``fair''.
The notion of fairness itself has been analysed at great depth in the literature on \emph{welfare economics} and \emph{distributive justice}. As we shall see, there have been a number of different proposals for turning the everyday notion of fairness into a precise mathematical definition. Each definition has its advantages and disadvantages, and understanding their exact properties is crucial before selecting any particular criterion to be used in an agent-based system or indeed any other type of application. We therefore review some of the most important fairness and efficiency criteria put forward in the literature and discuss their properties and axiomatic foundations.

\paragraph{Overview.}
These Notes are organised as follows.
We first cover some preliminaries, in particular terminology. 
Then, Section~\ref{sec:criteria} reviews a good number of fairness and efficiency criteria for assessing the economic quality of an allocation of resources. We emphasise criteria that can be formulated in terms of so-called social welfare orderings and provide a brief glimpse at their axiomatic foundations. We also cover Pareto efficiency, proportionality, envy-freeness, and degrees of envy.
Section~\ref{sec:cakes} is about cake-cutting procedures, as an example for fair division of a single divisible good (the cake).  We review some of the most important cake-cutting procedures for achieving proportional and envy-free divisions and discuss their properties.
Section~\ref{sec:indivisible} then introduces the problem of allocating a set of indivisible goods to a group of agents. We briefly comment on work aimed at establishing the computational complexity of different variants of this problem, and then discuss both centralised algorithms and a distributed approach based on negotiation.
Each of the three main sections concludes with some bibliographic notes, giving the most important references for the material presented and providing suggestions for further reading, as well as a small number of exercises.

\section{Preliminaries}

Throughout these notes, let $\mathcal{N}=\{1,\ldots,n\}$ be a finite set of \emph{agents} (often also referred to as \emph{individuals} or \emph{players}).
These agents need to agree on the division of a number of \emph{goods} (or \emph{resources}, \emph{items}, \emph{objects}, \emph{commodities}) between them.

\subsection{Types of Goods}

There are different \emph{types} of goods and the nature of a fair division problem will, to a large extent, depend on the type of goods under consideration.
Maybe the most important distinction is that between \emph{divisible} and \emph{indivisible} goods. An example for a divisible good would be a cake: you can allocate a slice of any size to an agent. An example for an indivisible good would be a book: it (arguably) only makes sense to allocate a book to an agent in one piece. Sometimes there are both divisible and indivisible goods; a typical example is the case of fair division with a set of indivisible goods and money (a divisible resource). 

Besides the distinction between divisible and indivisible goods, we may classify goods according to whether they are \emph{static} or whether they can potentially change their properties during the process of allocation (as is the case for \emph{perishable} or \emph{consumable} goods); whether they can or cannot be \emph{shared} between more than one agent; or whether they are available in \emph{single} or \emph{multiple units} (meaning that there may be several indistinguishable copies of the same good). The MARA Survey \citep{mara-survey} discusses these distinctions in some detail. In the present notes we will only consider fair division problems with static goods that are available in single units and that cannot be shared. We do consider both divisible and indivisible goods (the latter with and without monetary side-payments). 

\subsection{Allocations and Agreements}

Fair division tries to identify a desirable \emph{allocation} (or \emph{division}, \emph{assignment}) of goods to agents: for each agent, we need to specify which (part of) which item they should obtain. 

From a more abstract point of view, we can think of an allocation as an \emph{agreement} (or \emph{outcome}, \emph{solution}, \emph{alternative}, \emph{state of the world}). Many of the fairness criteria defined in Section~\ref{sec:criteria} are applicable beyond the sphere of resource allocation. To emphasise this fact we shall use the term \emph{agreement} whenever a concept is not specific to the domain of fair division but rather extends to any kind of collective decision between a group of agents. 

We model allocations as functions mapping agents to bundles of goods. The exact nature of the codomain depends on the details of the fair division model considered (see Sections~\ref{sec:cake-model} and~\ref{sec:indivisible-model}). We use letters $A$, $A'$, \ldots\ to denote allocations. For any agent $i\in\mathcal{N}$, $A(i)$ is the bundle given to~$i$ under allocation~$A$. We also use $A$, $A'$, \ldots\ to denote agreements of a more general nature.

\subsection{Preferences and Utility Functions}

Each agent $i\in\mathcal{N}$ is endowed with a \emph{utility function} $u_i$ mapping possible agreements/allocations to a suitable numerical scale. For the purposes of these Notes, suppose that this scale is always the set of real numbers.\footnote{For studies of the computational complexity of fair division problems (which we shall only briefly mention in Section~\ref{sec:complexity}) the rational numbers should be used instead.} 

\paragraph{Utility of a bundle vs.\ utility of an allocation.}
In the context of fair division, a very common assumption is that an agent's utility depends only on the goods that agent receives, rather than on goods received by some of the other agents or indeed any other aspect of the state of the world (\emph{``no externalities''}). That is, in actual fact, utility functions are typically defined over bundles of goods rather than over allocations. However, any such definition immediately extends to a definition of utility over allocations by simply stipulating that the utility of agent~$i$ in allocation~$A$ is $u_i(A(i))$, where $A(i)$ is the bundle of goods received by~$i$ under~$A$. That is, when utility functions are defined on bundles, then $u_i(A)$ is to be read as a shorthand for $u_i(A(i))$.

\paragraph{Utility vectors.}
Every feasible agreement $A$ induces a \emph{utility vector} $u(A)=\langle u_1(A),u_2(A),\ldots,u_n(A)\rangle$. If we order the elements of $u(A)$ in ascending order, then we obtain the \emph{ordered utility vector} of $A$, denoted by $u^*(A)$. We shall also speak about utility vectors $u=\langle u_1,\ldots,u_n\rangle\in\R^n$ and their ordered counterparts $u^*=\langle u^*_1,\ldots,u^*_n\rangle$ without always referring to the agreements inducing these vectors.

\begin{example}
Suppose we want to allocate four indivisible goods, say the set $\{a,b,c,d\}$, to three agents. We have to define a utility function $u_i:2^{\{a,b,c,d\}}\to\R$ from bundles of these goods to the reals for each agent $i=1,2,3$. Agent~1 only cares about the number of items she receives: $u_1:B\mapsto 10\cdot|B|$. Agent~2 gives utility $100$ to the full set: $u_2(\{a,b,c,d\})=100$. For any other set, she computes her utility by adding up the values she gives to the four individual items: $5$ to $a$, $23$ to $b$, $10$ to $c$, and $1$ to $d$ (that is, $u_2(\{a,b\})=28$, etc.). Agent~3 is not really interested and always happy. She assigns utility $25$ to every bundle she may receive, including the empty bundle: $u_3:B\mapsto 25$.
Now consider allocation $A$, which gives $c$ to agent~1, $a$ and $b$ to agent~2, and $d$ to agent~3. The resulting utility vector is $u(A)=\langle 10,28,25\rangle$ and the corresponding ordered utility vector is $u^*(A)=\langle 10,25,28\rangle$. While $u(A)$ tells us, amongst other things, that the \emph{first} agent enjoys utility~$10$, $u^*(A)$ tells us that the \emph{worst-off} agent enjoys utility~$10$.
If, instead of the allocation described, we give all four items to the first agent, then we obtain the utility vector $\langle 40,0,25\rangle$ and the ordered utility vector $\langle 0,25,40\rangle$.%
\end{example}

\paragraph{Utility and valuation functions.}
Note that in some parts of the literature the term \emph{valuation function} is preferred over the term \emph{utility function}. Furthermore, sometimes utility functions are defined in terms of valuation functions. For example, each agent may be endowed with a valuation function mapping bundles of goods to the reals, and an agent's utility could then be defined as the difference between the value assigned by that agent's valuation function to the bundle of goods obtained and the amount of money paid in return: \emph{utility} $=$ \emph{valuation} $-$ \emph{price} (so-called \emph{quasi-linear} utility function).

\paragraph{Utility functions and ordinal preference relations.}
From a cognitive point of view, a problematic aspect of using utility functions to model preferences is that they describe preferences in more detail than may be appropriate. For example, it may be quite reasonable to say that I like chocolate more than I like strawberries, which in turn I like more than potatoes ($C \succ S\succ P$), but not that the intensity of my preference of chocolate over potatoes is $2.5$ times as high as the intensity of my preference of strawberries over potatoes ($u(C)=20$, $u(S)=8$, $u(P)=2$). Similarly, it may not be reasonable to say that my appreciation of chocolate is higher than your appreciation of pumpkin pie. Despite these cautious remarks, using utility functions is very convenient from a purely technical point of view. We will therefore employ utility functions to model agent preferences throughout these Notes, but we will also, where appropriate, comment on which aspects of the expressive power of utility functions (including both \emph{preference intensity} and \emph{interpersonal comparison}) we actually require to be able to formulate certain concepts.

\section{Fairness and Efficiency Criteria}
\label{sec:criteria}

In this section, we shall give an overview of formal criteria for measuring and comparing the fairness, as well as the (economic) efficiency, of alternative allocations of goods. Most criteria (all except proportionality and envy-related concepts) can in fact be defined over arbitrary agreements and do not require us to speak about bundles of goods in particular.

\subsection{Pareto Efficiency}

The most fundamental efficiency criterion is the Pareto condition. An agreement is called \emph{Pareto efficient} (or \emph{Pareto optimal}) if there is no other feasible agreement that would make at least one agent strictly better off while not making any of the others worse off. 

\begin{definition}[Pareto dominance]
Agreement $A$ is Pareto dominated by agreement~$A'$ if $u_i(A)\leq u_i(A')$ for all agents $i\in\mathcal{N}$ and this inequality is strict in at least one case. 
\end{definition}

\begin{definition}[Pareto efficiency]
An agreement $A$ is Pareto efficient if there is no other feasible agreement $A'$ such that $A$ is Pareto dominated by~$A'$.
\end{definition}
Note that the definition of Pareto dominance (and thereby of Pareto efficiency) does not require us to be able to compare utilities across different agents, nor does it make any reference to the intensities of preferences. Indeed, simple preference orders (rather than fully fledged utility functions) would suffice to formulate the Pareto condition. 

Pareto efficiency is a very convincing criterion for judging the quality of of an agreement: if we can find another agreement that makes everyone at least as happy or even happier, we should probably go for that other agreement. On the other hand, Pareto efficiency is also a very weak criterion. For most fair division problems there will be \emph{many} alternative solutions that are all Pareto efficient (e.g., giving \emph{everything} to a single agent will be Pareto efficient whenever that agent has a strictly monotonic utility function). That is, Pareto efficient solutions need not be fair at all, and as a criterion Pareto efficiency is not very decisive. So, while we usually do want to satisfy (at least) Pareto efficiency, we need much stronger criteria to identify the truly interesting solutions.

\subsection{Collective Utility Functions and Social Welfare Orderings}
\label{sec:cufswo}

In principle, we may take all sorts of indicators into account when judging fairness. One particular stance to adopt would be to say that the only information we should use are the utility levels of the individual agents. This is known as the \emph{welfarist} approach. Technically, this means that, rather than looking at allocations (or, more generally speaking, agreements) and assessing their relative fairness, we only need to look at and compare the \emph{utility vectors} $\langle u_1,\ldots,u_n\rangle\in\R^n$ they induce. A whole range of fairness and efficiency criteria can be defined in terms of so-called social welfare orderings and collective utility functions.

\begin{definition}[Social welfare orderings]
A social welfare ordering (SWO) is a binary relation $\preceq$ over the space $\R^n$ of utility vectors that is reflexive, transitive and complete. 
\end{definition}
Intuitively, $u\preceq v$ expresses that the agreement inducing utility vector $v$ is socially at least as desirable as the agreement inducing utility vector $u$. 
We write $u\prec v$ (strict social preference) in case $u\preceq v$ but not $v\preceq u$; and we write $u\sim v$ (social indifference) in case both $u\preceq v$ and~$v\preceq u$.

Most social welfare orderings can be defined in terms of a collective utility function.

\begin{definition}[Collective utility functions]
A collective utility function (CUF) is a function $\sw{}:\R^n\to\R$ mapping utility vectors to the reals. 
\end{definition}
The collective utility $\sw{}(u)$ of a utility vector $u$ is often called the \emph{social welfare} of that vector (and of the agreement/allocation that induces that vector). Every CUF \emph{induces} an SWO: $u\preceq v$ if and only if $\sw{}(u)\leq \sw{}(v)$. 

\paragraph{Important collective utility functions.}
We now define the most important CUFs.
The first is the utilitarian CUF, which defines the social welfare of an agreement as the sum of utilities it generates in the individual agents. It is a pure efficiency criterion.

\begin{definition}[Utilitarian social welfare]
The utilitarian CUF is mapping each utility vector to the sum of individual utilities:
\begin{eqnarray*}
\sw{util}(u) & = & \sum_{i\in\mathcal{N}} u_i
\end{eqnarray*}
\end{definition}
Again, while CUFs are formally defined over utility vectors, they also induce a corresponding notion of social welfare of agreements. In the case of the utilitarian CUF, for example, we have $\sw{util}(A)=\sw{util}(\langle u_1(A),\ldots,u_n(A)\rangle)=\sum_{i\in\mathcal{N}}u_i(A)$ for any given agreement~$A$.

An agreement with maximal utilitarian social welfare is an agreement that maximises average utility, which explains why this may be considered an attractive social criterion. On the other hand, this definition of social welfare completely ignore fairness considerations: an allocation giving utility 101 to one agent and 0 to another would be considered socially superior to an allocation giving both of them utility~50.

A very different notion of social welfare is epitomised by the egalitarian CUF, which identifies social welfare with the utility level of the worst-off member of society.

\begin{definition}[Egalitarian social welfare]
The egalitarian CUF is mapping each utility vector to the minimum individual utility:
\begin{eqnarray*}
\sw{egal}(u) & = & \min\{u_i \mid i\in\mathcal{N}\}
\end{eqnarray*}
\end{definition}
That is, maximising egalitarian social welfare amounts to raising the utility of the worst-off member of society (whoever that may end up being) as much as possible. 

If we take the maximum rather than the minimum utility as an indicator, we obtain an elitist form of social welfare.
\begin{definition}[Elitist social welfare]
The elitist CUF is mapping each utility vector to the maximum individual utility:
\begin{eqnarray*}
\sw{elit}(u) & = & \max\{u_i \mid i\in\mathcal{N}\}
\end{eqnarray*}
\end{definition}
Clearly, elitist social welfare has little in common with any intuitive notion of fairness. In an application in which all that matters is that at least one agent achieves their goal, it may however be the perfect formalisation of the social desirability of a state.

The elitist and egalitarian CUF are both instances of the class of $k$-rank dictator CUFs.

\begin{definition}[$k$-rank dictators]
Let $k\leq n$. The $k$-rank dictator CUF is mapping each utility vector to the $k$th element of the corresponding ordered vector:
\begin{eqnarray*}
\sw{{\it k}}(u) & = & u^*_k
\end{eqnarray*}
\end{definition}
For $k=1$ we obtain the egalitarian CUF, and for $k=n$ we obtain the elitist CUF. A third special case of particular interest is the \emph{median rank dictator} CUF with $k=\lfloor \frac{n+1}{2}\rfloor$. For utilities drawn from a uniform distribution it will make similar social welfare judgements as the utilitarian CUF, but it is ``blind'' with respect to agents that are either extremely well or extremely badly off.

Another important notion of social welfare is inspired by the Nash bargaining solution.
\begin{definition}[Nash social welfare]
The Nash CUF (also known as the Nash product) is mapping each utility vector to the product of individual utilities:
\begin{eqnarray*}
\sw{nash}(u) & = & \prod_{i\in\mathcal{N}} u_i
\end{eqnarray*}
\end{definition}
The Nash CUF combines efficiency and fairness considerations. Like the utilitarian CUF it favours high total utility. But at the same time it also encourages inequality-reducing transfers of utility. For example, the utilitarian CUF cannot distinguish between $\langle 4,4\rangle$ and $\langle 2,6\rangle$, while the Nash CUF will favour the former.

\paragraph{Leximin ordering.}
Recall the definition of egalitarian social welfare, which stipulates that striving for fairness means maximising the utility of the worst-off member of society. A natural refinement of this idea is this: first maximise the utility of the worst-off, and once all possibilities for improving minimum utility have been exhausted, maximise the second worst-off utility as much as possible, and then the third, and so forth. 
This idea has been formalised as the so-called leximin ordering.

\begin{definition}[Leximin ordering]
The leximin ordering $\preceqlex$ is the SWO which for any two utility vectors $u$ and $v$ (with corresponding ordered vectors $u^*$ and $v^*$) stipulates $u\preceqlex v$ if and only if $u^*=v^*$ or there exists a $k\leq n$ such that $u^*_k<v^*_k$ and $u^*_j=v^*_j$ for all $j<k$.
\end{definition}
In other words, $u\preceqlex v$ exactly when $u^*$ lexicographically precedes $v^*$ (or when they are the same). Note that $u\preceqlex v$ entails $\sw{egal}(u)\leq \sw{egal}(v)$ (and, equivalently, $\sw{egal}(u)<\sw{egal}(v)$ entails $u\preclex v$, for the strict variant of the leximin ordering), but not \textit{vice versa}.


The leximin ordering $\preceqlex$ is an example for an SWO that \emph{cannot} be represented by a CUF \citep[Lemma~2.1]{Moulin1988}.
It can, however, be approximated, as we shall see next.

\paragraph{Ordered weighted averaging operators.}
Several of the criteria introduced so far may be regarded as particular instances of ordered weighted averaging operators.

\begin{definition}[Ordered weighted averaging]
Let $w\in\R^n$. The ordered weighted averaging operator parametrised by $w$ is the CUF mapping each utility vector $u$ to:
\begin{eqnarray*}
\OWA_w(u) & = & \sum_{i\in\mathcal{N}} w_i\cdot u^*_i
\end{eqnarray*}
\end{definition}
That is, we multiply the $i$th element of the ordered variant of $u$ with weight $w_i$, for all $i\in\mathcal{N}$, and add up the results.
Observe that $\OWA_w$ for $w=\langle 1,1,\ldots,1\rangle$ is just the utilitarian CUF. For $w_k=1$ and $w_i=0$ whenever $i\not= k$, $\OWA_w$ coincides with the $k$-rank dictator CUF. If $w_i=\alpha^{i-1}$ with $\alpha>0$, then the SWO induced by $\OWA_w$ converges to the leximin ordering as we let $\alpha$ go to~$0$.

\paragraph{Axiomatic approach.}
We have mentioned a few intuitive arguments in support of the choice of one particular SWO over another. Intuitions are important, but should be made as precise as possible. If we can formulate formal definitions, or \emph{axioms}, of the properties that we would like our SWO of choice to have, then we can make clear statements about which SWOs do and do not satisfy a given combinations of such desiderata. Besides checking whether a given SWO satisfies a given axiom, in some cases it is also possible to fully characterise an SWO (or a class of SWOs) by means of a set of axioms. Such a result tells us that the SWO at hand is the \emph{only} fairness criterion that meets our requirements. Similarly, for some combinations of axioms it will be possible to prove that there can be no SWO that would satisfy all of them. This approach is known as the \emph{axiomatic method} in social choice theory and welfare economics. We will review some of the most important axioms here and mention a number of characterisation results (for precise statements of these results, please refer to the cited literature; cf.\ Section~\ref{sec:criteria:literature}).

A very basic axiom is anonymity. It states that all agents should be treated equally, in the sense that their position in the ordering imposed on agents does not affect social welfare judgements.

\begin{axiom}[Anonymity]
An SWO $\preceq$ is anonymous if $u$ being a permutation of $v$ entails $u\sim v$ for all $u,v\in\R^n$.
\end{axiom}
Anonymity has also been called \emph{symmetry}. 
Clearly, all SWOs we have presented satisfy anonymity (indeed, anonymity is sometimes considered part of the definition of what makes an SWO).

Another important axiom is unanimity. It states that if all agents (weakly) prefer one agreement over another, then so should society.
\begin{axiom}[Unanimity]
An SWO $\preceq$ is unanimous if $(i)$~$u\preceq v$ holds whenever $u_i\leq v_i$ for all agents $i\in\mathcal{N}$ and $(ii)$~$u\prec v$ holds whenever $u_i < v_i$ for all agents $i\in\mathcal{N}$.
\end{axiom}
Note that unanimity is closely related to Pareto efficiency. 
All of the SWOs presented here satisfy unanimity.
(However, some do not satisfy the Pareto condition, in the sense that an agreement that is optimal according to, say, the elitist SWO need not be Pareto efficient as well---but there are always agreements that are both Pareto efficient and optimal with respect to the elitist SWO.)

A further basic requirement is that social welfare judgements should be independent of non-concerned agents. This condition is also known as separability.

\begin{axiom}[Separability]
An SWO $\preceq$ is separable if 
$u\preceq v$ entails $(u+w)\preceq(v+w)$
for all $u,v,w\in\R^n$ with $w_i=0$ whenever $u_i\not=v_i$.
\end{axiom}
For example, suppose we rank $u$ above $v$ and, say, agent~$i$ enjoys the same utility for both of them. Then separability says that if we uniformly change the utility level of~$i$ for both $u$ and $v$, then we will still rank the (changed) $u$ above the (changed) $v$.
The utilitarian SWO and the leximin ordering are examples for SWOs that satisfy this axiom.
The egalitarian SWO, on the other hand, violates separability: $\langle 1,7,8\rangle \preceq_{\scriptsize{\rm egal}} \langle 1,3,5\rangle$ (indeed, those two utility vectors are equally preferred under egalitarianism) but $\langle 1+10,7,8\rangle \not\preceq_{\scriptsize{\rm egal}} \langle 1+10,3,5\rangle$. In fact, no $k$-rank dictator SWO satisfies separability. In other words, this axiom is more demanding that it may appear at first glance, and we may not always want to impose it.

Now that we have some basic axioms in place that allow us to narrow down the (very large) space of SWOs (i.e., the space of reflexive, transitive and complete relations over $\R^n$) to a range of reasonable definitions, we can turn our attention to additional axioms that identify specific (fairness) properties that we may or may not want to impose. 
Our first axiom of this kind, the Pigou-Dalton principle, encapsulates a central intuition about fairness: a fair SWO should encourage inequality-reducing redistributions of welfare. 

\begin{axiom}[Pigou-Dalton principle]
An SWO $\preceq$ respects the Pigou-Dalton principle 
if, for all $u,v\in\R^n$, $u\preceq v$ holds whenever there
exist $i,j\in \mathcal{N}$ such that: 
\begin{itemize}
\item $u_k=v_k$ for all $k\in \mathcal{N}\setminus\!\{i,j\}$
--- only $i$ and $j$ are involved;
\item $u_i+u_j=v_i+v_j$
--- the change is mean-preserving; and
\item $|u_i-u_j|>|v_i-v_j|$
--- the change is inequality-reducing.
\end{itemize} 
\end{axiom}
The leximin ordering, for example, satisfies the Pigou-Dalton principle.
So does the utilitarian SWO, but only in a very weak sense: it is in fact ``blind'' towards inequality-reducing redistributions as long as they are mean-preserving. 

Observe that each of the various SWOs introduced relies on cardinal utility information; that is, ordinal preferences alone would not suffice to state these definitions. We need to be able to make utility comparisons across agents and the intensity of preferences does matter. However, not every SWO uses \emph{all} of the information present in the utility vectors when we use that SWO to compare two vectors. For instance, if we consistently multiply the utility of agent~7 with some constant factor $c$ for every possible agreement, then this will not affect social welfare judgements when the Nash SWO is being used. That is, the absolute values of utilities do not matter in this case, only the ratio between the utilities enjoyed by an agent for different agreements does. The same is not true for the egalitarian SWO for instance. In that case, if agent~7 happens to be the worst-off agent for one of two given agreements, then multiplying her utility with $c$ in both cases may mean that a different agent will end up being worst-off and the social welfare judgement could be reversed. 
Such differences can be captured formally, by means of suitable axioms. The remaining axioms are all of this kind.

Suppose we compare the utility vectors induced by two allocations before and after a deal.
If agents did enjoy very different levels of utility before the encounter,
it may not be meaningful to use their absolute utilities afterwards to 
assess social welfare, but we should maybe rather refer to their relative gain or loss in utility.
So a desirable property of an SWO may be to be independent of
what individual agents consider ``zero'' utility.

\begin{axiom}[ZI]
An SWO $\preceq$ is zero independent if
$u\preceq v$ entails $(u+w)\preceq(v+w)$ for all $u,v,w\in\R^n$. 
\end{axiom}
For example, the utilitarian SWO is zero independent, while the egalitarian SWO and the Nash SWO are not. 
In fact, under some mild technical assumptions (that are, for instance, satisfied if the set of feasible agreements is finite), amongst all the SWOs that are anonymous, unanimous, separable, and that satisfy the Pigou-Dalton principle, the utilitarian SWO is the \emph{only} SWO that satisfies~ZI. In this sense, the utilitarian SWO is (essentially) axiomatised by the axiom of zero independence.

Different agents may measure their personal utility using different
``currencies'', such as dollars as opposed to euros. 
So a desirable property of an SWO may be to be independent 
of the utility scales used by individual agents.
For the next axiom, we assume that all utilities are positive, i.e., $u\in(\R^+)^n$.
Let $u\cdot v=\langle u_1\cdot v_1,\ldots,u_n\cdot v_n\rangle$.

\begin{axiom}[SI]
An SWO $\preceq$ over positive utilities 
is scale independent if
$u\preceq v$ entails $(u\cdot w)\preceq(v\cdot w)$ for all $u,v,w\in(\R^+)^n$. 
\end{axiom}
Clearly, neither the utilitarian nor the egalitarian SWO are scale independent.
The Nash product, on the other hand, does satisfy~SI. 
In fact, SI characterises the Nash SWO in the same way as ZI characterises the utilitarian SWO.

If we weaken zero independence and scale independence to require that social welfare judgements should only be independent of a common change of zero and a common change of scale, respectively, we obtain two further important axioms.
Let $e=\langle 1,\ldots,1\rangle\in\R^n$.

\begin{axiom}[ICZ]
An SWO $\preceq$ is independent of the common zero of utility 
iff $u\preceq v$ entails $(u+\lambda e)\preceq(v+\lambda e)$ for all
$u,v\in\R^n$ and all $\lambda\in\R$.
\end{axiom}

\begin{axiom}[ICS]
An SWO $\preceq$ over positive utilities 
is independent of the common utility scale if
$u\preceq v$ entails $\lambda u\preceq\lambda v$
for all $u,v\in(\R^+)^n$ and all $\lambda\in\R^+$.
\end{axiom}
To illustrate another desirable property that we may wish our SWO to satisfy, consider the following example. 
Suppose we would like to be able to make social welfare judgements without knowing what kind of tax members of society will have to pay. 
Think of tax as a function $f$ that maps gross income to net income after taxes. We may not know the precise tax rule, but we do know (at least under some idealising assumptions) that higher gross income will result in higher net benefit.  
\begin{axiom}[ICP]
An SWO $\preceq$ is independent of the common utility pace
if $u\preceq v$ entails $f(u)\preceq f(v)$ for all $u,v\in\R^n$ 
and for every increasing bijection $f:\R\to\R$.
\end{axiom}
For an SWO satisfying ICP only ordinal comparisons
($u_i\leq v_j$ or $u_i\geq v_j$) matter, but the (cardinal)
intensities $u_i-v_j$ do not.
The utilitarian SWO is not independent of the common utility pace,
but the egalitarian SWO is.
In fact, any $k$-rank dictator SWO is.

\subsection{Proportionality}

Suppose agents have monotonic utility functions declared over bundles of goods (a utility function $u$ is monotonic if $B\subseteq B'$ entails $u(B)\leq u(B')$). Then an agent will be happiest if they receive \emph{all} the goods, and they may feel entitled to at least $1/n$ of the value of this full bundle. The criterion of proportionality is satisfied when each agent believes that they received a fair share of the goods.

\begin{definition}[Proportionality]
An allocation $A$ is proportional if $u_i(A(i))\geq \frac{1}{n}\cdot\hat{u}_i$ for every agent $i\in\mathcal{N}$, where $\hat{u}_i$ is the utility given to the full bundle by agent~$i$. 
\end{definition}
Proportionality is considered an attractive fairness criterion when utility functions are additive (that is, when the utility of a bundle can always be computed as the sum of the utilities of the parts of that bundle, whichever way we choose to partition the bundle). If utility functions are subadditive (that is, when the sum of utilities of two disjoint bundles will be less or at most equal to the utility assigned to their union), then proportionality may be too easy to satisfy to be of any real interest. Conversely, when utility functions are superadditve, then it may often be impossible to satisfy proportionality.
 
\subsection{Envy-freeness and Degrees of Envy}

An allocation, assigning a bundle of goods to each agent, is called envy-free if no agent strictly prefers one of the bundles assigned to another agent to their own bundle. 

\begin{definition}[Envy-freeness]
An allocation $A$ is envy-free if $u_i(A(i))\geq u_i(A(j))$ for every pair of agents $i,j\in\mathcal{N}$.
\end{definition}
An attractive aspect of envy-freeness as a fairness criterion is that it can be defined in terms of ordinal preference information alone.

If we require allocations to be \emph{complete} (that is, every good needs to be allocated to some agent), then envy-free allocation do not exist for some combinations of utility functions. For example, if there are two agents and just one item and that item has positive utility for both agents, then neither one of the two possible complete allocations is envy-free (but the incomplete allocation that assigns the empty bundle to both of them is, vacuously, envy-free).

\paragraph{Degrees of envy.} 
As eliminating envy entirely is not always an option, we may instead seek to reduce envy as much as possible. This presupposes a formal measure for the degree of envy associated with an allocation. We can define several such measures using a three-stage process. First, we need to define the degree of envy experienced by one agent towards another (single) agent. Once such a definition is in place, we can aggregate over all agents envied (or not) by a particular agent and define the degree of envy of that agent towards the rest of society. Aggregating one more time, we can eventually define what should be understood by the degree of envy experienced by society as a whole given a particular allocation of goods. At each of the three stages we have several options, some of which are listed below:
\begin{itemize}
\item\emph{Envy between pairs of agents.}
The degree to which agent~$i$ envies agent~$j$ under allocation~$A$ may be measured in a number of ways.
\begin{itemize}
\item Positive: $\max\{u_i(A(j))-u_i(A(i)),0\}$ ---
the difference between the utility assigned to $j$'s bundle and her own bundle if~$i$ prefers the other bundle, and $0$ otherwise
\item Total: $u_i(A(j))-u_i(A(i))$ ---
also allowing for ``negative envy'', in case~$i$ feels she is actually better off than~$j$
\item Boolean: $1$ if $u_i(A(j))>u_i(A(i))$ and $0$ otherwise 
\end{itemize}
\item\emph{Envy of a single agent.}
To compute the envy of agent~$i$ towards the rest of society, we need to aggregate $i$'s envies towards all other agents (computed using any of the three measures above).
\begin{itemize}
\item Sum: take the sum of envies towards other agents (similar to the utilitarian CUF)
\item Maximum: take the maximum (similar to the egalitarian CUF)
\end{itemize}
In principle, aggregation operators derived from other CUFs may also be considered.
\item\emph{Envy of society.}
To compute the degree of envy experienced by society we need to aggregate the individual envies that each agent experiences towards the others.
\begin{itemize}
\item Sum: take the sum of individual envies (utilitarian)
\item Maximum: take the maximum of individual envies (egalitarian)
\end{itemize}
Again, additional forms of aggregation may also be of interest.
\end{itemize}
For instance, using the combination Boolean/Maximum/Sum amounts to counting the number of envious agents in society.

\subsection{Bibliographic Notes}
\label{sec:criteria:literature}

\citet{Moulin1988} offers an excellent introduction to the theory of social welfare orderings and discusses its axiomatic foundations in detail. (This has been the main reference for the preparation of Section~\ref{sec:cufswo}.) For a less technical introduction covering similar grounds, consult the undergraduate textbook by the same author \citep{Moulin2003}. For further reading, the books by \citet{Sen1970} and \citet{Roemer1996} are highly recommended works (of a technical nature) on welfare economics and distributive justice. 
\citet{ThomsonRCER539} discusses the axiomatic foundations of the envy-freeness criterion as well as other ordinal fairness criteria (as opposed to those based on cardinal utility information).
Our discussion of degrees of envy follows \citet{ChevaleyreEtAlIJCAI2007}.

An important class of fairness criteria that we have \emph{not} covered are those specifically aimed at measuring the degree of economic \emph{inequality} generated by an agreement. These criteria include, amongst others, the \emph{Lorenz curve} and the \emph{Gini index} \citep{AtkinsonJET1970,Sen1973,Shorrocks1988}.

The most widely used criteria for assessing the economic qualities of an allocation of resources in the early literature on multiagent systems have been Pareto efficiency and utilitarian social welfare. Some of the arguments for considering instead the whole range of criteria proposed in the welfare economics literature (and for developing further such criteria, including tailor-made criteria for specific applications) are spelt out elsewhere \citep{EndrissMaudetESAW2003,mara-survey}. 

\subsection{Exercises}

\begin{exercise}
Which of the following statements is true? 
Give either a proof (in the affirmative case) or a counterexample (otherwise).
\begin{enumerate}
\item Any agreement with maximal utilitarian social welfare is Pareto efficient.
\item No agreement can maximise both utilitarian and egalitarian social welfare.
\item Any agreement that is optimal with respect to the leximin ordering is both Pareto efficient and maximises egalitarian social welfare.
\item If preferences are dichotomous (meaning: $u_i(A)=0$ or $u_i(A)=1$ for any agent~$i$ and any agreement~$A$), then the utilitarian SWO and the leximin ordering coincide. 
\item The egalitarian SWO respects the Pigou-Dalton transfer principle, and it is the only $k$-rank dictator SWO to do so.
\end{enumerate}
\end{exercise}

\begin{exercise}
If there are only two agents and preferences are additive, then envy-freeness and proportionality express the same property: 
\begin{itemize}
\item Any envy-free allocation is also proportional.
\item Any proportional allocation is also envy-free.
\end{itemize}
Check to what extent these two statements remain true when we relax the assumptions. Regarding the number of agents, consider the case of two and the case of an arbitrary number; regarding the preferences, consider additive, subadditive, superadditive, and arbitrary preferences.
(Beware that most of the statements regarding proportionality and envy-freeness that you may find in the literature refer to the case of additive preferences only.)
\end{exercise}

\begin{exercise}
Suppose there are $n$ agents located anywhere on the interval $[0,1]$.
We have to decide where to build an amusement park $A$, also anywhere
on the same interval. The \emph{dis}utility of an agent is its distance to $A$.
\begin{enumerate}
\item What is the solution selected by the egalitarian CUF?
\item What is the solution selected by the elitist ($n$-rank dictator) CUF?
\item For arbitrary $k\leq n$, give a general algorithm to compute a solution
that is optimal with respect to the $k$-rank dictator CUF.
What is the complexity of your algorithm?
\end{enumerate}
\end{exercise}

\section{Divisible Goods: Cake-Cutting Procedures}
\label{sec:cakes}

In this section, we shall be concerned with the problem of fair division for divisible goods, or more precisely a single divisible good. An example for such a divisible good is a cake. The main fairness criteria that have been considered in the cake-cutting literature are proportionality and envy-freeness.

We first discuss the case of $n=2$ agents between which to divide the cake. As we shall see, there is a very simple procedure that can guarantee both proportionality and envy-freeness. For $n>2$ the problem of designing a fair division procedure is considerably more challenging. We will present several procedures that can guarantee proportional outcomes with arbitrary numbers of agents, but we will also see that there are no such simple and general solutions for the envy-free case. Instead, we will present two envy-free procedures for $n=3$ agents and briefly comment on the difficulty of the general problem of envy-free cake-cutting.

\subsection{The Model}
\label{sec:cake-model}

Let $\mathcal{N}=\{1,\ldots,n\}$ be a set of $n$ \emph{agents} (often referred to as \emph{players} in the cake-cutting literature).
We will be particularly interested in the cases of $n=2$ and $n=3$.
These agents need to divide a \emph{cake} amongst themselves by means of a series of parallel cuts. The cake is represented by the unit interval $[0,1]$:
\begin{center}
\verb+|----------------------|+ \\
\verb+0                      1+
\end{center}
A \emph{bundle} will be a finite union of subintervals of the full cake. These subintervals are not allowed to overlap (so goods cannot be shared) and we shall only be interested in complete allocations, where every piece of the cake is allocated to someone. 

Each agent $i\in\mathcal{N}$ has got a \emph{utility function} $u_i$ (also referred to as the agent's \emph{valuation} or \emph{measure}) mapping finite unions of subintervals of $[0,1]$ to the reals, that satisfies the following conditions:
\begin{itemize}
\item Non-negativity: $u_i(B)\geq 0$ for all $B\subseteq [0,1]$
\item Normalisation: $u_i(\emptyset)=0$ and $u_i([0,1])=1$
\item Additivity: $u_i(B\cup B')=u_i(B)+u_i(B')$ for disjoint $B,B'\subseteq [0,1]$
\item $u_i$ is continuous: the Intermediate-Value Theorem applies and single points do not have any value. Specifically, if $0<x<y\leq 1$ with $u_i([0,x])=\alpha$ and $u_i([0,y])=\beta$, then for every $\gamma\in[\alpha,\beta]$ there exists a $z\in[x,y]$ such that $u_i([0,z])=\gamma$.
\end{itemize}
It is common (and reasonable) to strengthen the non-negativity assumption to require that $u_i(B)>0$ for all \emph{nonempty} $B\subseteq [0,1]$. That is, under this stronger assumption every player will assign \emph{some} positive value to every (proper) slice of cake. 

\subsection{Cut-and-Choose}

Suppose there are only $n=2$ players. 
There is a very simple and well-known procedure that we can use in this case.

\begin{cakeproc}{Cut-and-Choose}{2}
One player \emph{cuts} the cake in two pieces (which she considers
to be of equal value), and the other one \emph{chooses} one of the pieces
(the piece she prefers).
\end{cakeproc}
In the description of the procedure, the parts shown in parentheses are strictly speaking not part of the procedure, but rather spell out the strategy that each player should follow. It is not hard to see that if a player follows the recommended strategy, then they can \emph{guarantee} a proportional piece for themselves (at least half of the cake, according to their own utility function), \emph{whatever} the other agent is doing.\footnote{In fact, the first player (assuming she is risk-averse and will play so as to guarantee her a fair share whatever the other player may do) will receive exactly $1/2$, while the second will usually get more.}
That is, cut-and-choose is a proportional procedure for two players. It is also an envy-free procedure, because for $n=2$ (and additive utility functions) the two concepts do in fact coincide. 

Whether or not we want to say that cut-and-choose guarantees \emph{Pareto efficient} outcomes depends on what we consider the space of feasible divisions. Amongst all divisions using only a single cut, cut-and-choose clearly does guarantee Pareto efficiency (assuming players never give utility~0 to a proper slice). 
If we also consider divisions with more than one cut feasible, then cut-and-choose does not ensure Pareto efficiency. For example, if the first agent is very keen on the middle part of the cake and the second agent has high utility for both the part on the very left of the cake and the part of its very right, then a Pareto efficient division would require (at least) two cuts. This latter definition of Pareto efficiency is, arguably, too demanding. In particular, it is not difficult to see that no procedure using a bounded number of cuts could ever guarantee this form of Pareto efficiency, even for the simple case of just two players. 

\subsection{Operational Properties}

Besides fairness and efficiency properties, we can also evaluate cake-cutting procedures according to some other types of properties:
\begin{itemize}
\item Does the procedure guarantee that each agent receives a single 
\emph{contiguous} slice (rather than the union of several subintervals)?
If possible, we prefer such contiguous procedures, which also minimise the number of cuts that need to be made. Note that a procedure for $n$ players will require at least $n{-}1$ cuts.
\item If the number of cuts is not minimal, can we at least provide an \emph{upper bound} on the number of cuts? Clearly, the lower the number of cuts, the better. (There are procedures where no such bound can be given \textit{a~priori}.)
\item What is the \emph{complexity} of the procedure, measured in terms of the number of basic operations? Such basic operations are often taken to be two types of queries: $(i)$~asking an agent~$i$ to indicate a point~$x$ on the cake such that $u_i([0,x])=\alpha$ for a given value~$\alpha$; and $(ii)$~asking an agent~$i$ to specify their utility for a given slice~$[x,y]\subseteq [0,1]$.
\item Does the procedure require an active \emph{referee}, or can all actions 
be performed by the players themselves?
\item Is the procedure an algorithm in the proper sense of the word (also known as a \emph{protocol})? That is, can it be implemented by means of a clearly defined sequence of queries to the agents? (As we shall see there are some procedures, so-called moving-knife procedures, that cannot be translated into a discrete sequence of steps.) 
\end{itemize}
Cut-and-choose is ideal and as simple as can be with respect to all of these properties.

\subsection{Proportional Procedures for $n>2$}

We now review cake-cutting procedures for more than two agents when the goal is to ensure proportionality of the outcome. Note that when there are more than two agents, then proportionality is a less demanding criterion than envy-freeness, which is why we deal with proportionality first. Historically, the first cake-cutting procedures are those of Steinhaus and of Banach and Knaster (first reported in 1948). 

\paragraph{Steinhaus procedure.}
The Steinhaus procedure can be applied to cake-cutting problems with $n=3$ players. 

\begin{cakeproc}{Steinhaus Procedure}{3}
\begin{enumerate}
\item Player~1 cuts the cake into three pieces (which she values equally).
\item Player~2 ``passes'' (if she thinks at least two of the pieces are $\geq 1/3$)
or labels two of them as ``bad''. ---
If player~2 passed, then players~3, 2, 1 each choose a piece
(in that order) and we are done. $\checkmark$
\item If player~2 did not pass, then player~3 can also choose between passing
and labelling. ---
If player~3 passed, then players 2, 3, 1 each choose a piece (in that order)
and we are done. $\checkmark$
\item If neither player~2 or player~3 passed, then player~1 has to take (one of) 
the piece(s) labelled as ``bad'' by both 2 and 3. --- 
The rest is reassembled and 2 and 3 play cut-and-choose. $\checkmark$ 
\end{enumerate}\ \\[-35pt]
\end{cakeproc}
The Steinhaus procedure ensure proportional outcomes, but it does not guarantee envy-freeness. The (maximum) number of cuts is not minimal (3 rather than~2), and the pieces will not always be contiguous. The procedure does not require the active participation of a referee and it is an algorithm in the strong sense of the word.

\paragraph{Banach-Knaster procedure.}
The Banach-Knaster procedure, also known as the last-diminisher procedure, works for any number of players.

\begin{cakeproc}{Banach-Knaster Last-Diminisher Procedure}{$n$}
\begin{enumerate}
\item Player~1 cuts off a piece (that she considers to represent $1/n$).
\item That piece is passed around the players. Each player either lets it pass
(if she considers it too small) or trims it down further (to what she considers $1/n$).
\item After the piece has made the full round, the last player to cut something off
(the ``last diminisher'') is obliged to take it.
\item The rest (including the trimmings) is then divided amongst the remaining $n{-}1$ players.
Play cut-and-choose once $n=2$. $\checkmark$ 
\end{enumerate}\ \\[-35pt]
\end{cakeproc}
The procedure's properties are similar to those of the Steinhaus procedure: it guarantees proportional (but not envy-free) outcomes, the number of cuts is bounded, and no external referee is required. The resulting division need not be contiguous, but we can turn the Banach-Knaster procedure into a contiguous procedure by means of a small refinement: if slices are cut from the left of the cake, only allow agents to trim slices from the right. Another way of looking at this would be that the agent cutting off a new piece does in fact only \emph{indicate} where she would cut the cake, and rather than trimming the piece, subsequent agents only move the knife further to the left (we only actually cut the cake when it is clear who will have to take the piece in question).

\paragraph{Dubins-Spanier procedure.}
In 1961, Dubins and Spanier proposed a so-called \emph{moving-knife procedure} for an arbitrary number of agents. 
\begin{cakeproc}{Dubins-Spanier Procedure}{$n$}
\begin{enumerate}
\item A referee moves a knife slowly across the cake, from left to right.
Any player may shout ``stop'' at any time. Whoever does so receives the piece
to the left of the knife.
\item When a piece has been cut off, we continue with the remaining $n{-}1$
players, until just one player is left (who takes the rest).~$\checkmark$
\end{enumerate}\ \\[-35pt]
\end{cakeproc}
The Dubins-Spanier guarantees proportionality, but not envy-freeness. Observe that the agent that takes the first piece will obtain exactly $1/n$ (assuming she is risk-averse and is playing her security strategy of shouting ``stop'' as soon as the piece to the left of the knife represents a fair share for her), while the agent taking the last piece is best off (this is the same for the Banach-Knaster procedure). The procedure produces contiguous slices (and hence requires a minimal number of actual cuts). It requires the active help of a referee (namely to move the knife). 

Importantly, it is \emph{not} an algorithm in the true sense of the word. The problem is that it is not possible to construct an agent that would be able to continuously monitor the moving knife and to react at the exact moment when the abstract description of their perfect strategy would require them to do so. Of course, we \emph{can} approximate the ideal abstract situation as closely as we wish, and for utility functions that are sufficiently ``smooth'' there may be no problem in practice. Also, the problem only arises if we insist to faithfully model the moving-knife approach; if we instead ask agents to indicate at which point they \emph{would} shout ``stop'' if the knife \emph{were} to make it to that point, then we can easily discretise the procedure (and end up with something very similar to the Banach-Knaster procedure).

\paragraph{Even-Paz procedure: complexity.}
In 1984, Even and Paz proposed a procedure, also known as the divide-and-conquer procedure, specifically for the purposes of studying the complexity of procedures that guarantee proportionality.

\begin{cakeproc}{Even-Paz Divide-and-Conquer Procedure}{$n$}
\begin{enumerate}
\item Ask each player to indicate her 
$\lfloor\frac{n}{2}\rfloor$ / $\lceil\frac{n}{2}\rceil$ mark.
\item Associate the part to the left of the $\lfloor\frac{n}{2}\rfloor$th mark
with the players who made the leftmost $\lfloor\frac{n}{2}\rfloor$ marks
(group~1), and the rest with the others (group~2).
\item Recursively apply the same procedure to each of the two groups,
until only a single player is left. $\checkmark$
\end{enumerate}\ \\[-35pt]
\end{cakeproc}
For example, if $n=7$, then in step (1) each agent is asked to indicate a point $x$ on the cake such that $[0,x]$ has value $3/7$ and $[x,1]$ has value $4/7$. Then we cut the cake at the 3rd mark and associate the 3 agents who made the 3 leftmost marks with the left part of the cake, and the remaining agents with the right part of the cake. This kind of local procedure is iterated for each part, until we are down to the level of individual agents. It is not hard to check that the Even-Paz procedure requires $O(n\log n)$ marking queries until it terminates with a proportional division of the cake (see Exercise~\ref{ex:even-paz}). For large values of $n$, no procedure can guarantee a proportional outcome using fewer queries. 

The Even-Paz procedure guarantees proportionality (but not envy-freeness), produces contiguous pieces, and does not require the assistance of an external referee.

\subsection{Envy-free Procedures for $n>2$}

Devising a cake-cutting procedure that will guarantee envy-free outcomes is considerably more difficult than guaranteeing proportionality alone. 

The reason why none of the procedures for arbitrary numbers of agents described above can guarantee envy-freeness can be condensed into a simple argument. For all of them, agents get assigned their pieces one-by-one, and once an agent has received their piece they can no longer influence the division of the remaining cake amongst the remaining agents. So even if the agent has received what she perceives as a fair share, as she has no control over the remaining steps in the procedure, and as the other agents' utility functions may be very different from her own, whenever the remaining piece of the cake is more valuable to her than her own assigned piece, it is possible that the remaining agents will decide to divide that piece in a way that she perceives as very unfair, leaving one of them with a piece better than her own.

In fact, at the time of writing, no entirely satisfying envy-free procedure has been found (and this situation is unlikely to change in the near future; there are even reasons to believe that it may be impossible to devise such a procedure). In particular, there is no known procedure for $n\geq 5$ agents that can guarantee an envy-free division using a number of cuts bounded by any function of~$n$. Even for $n=4$ agents, there is no known procedure producing an envy-free division with contiguous pieces.

In this section, we shall present the two classic procedures for envy-free cake-cutting with $n=3$ players. Neither of them is perfect: one of them will not always assign contiguous slices of cake to the players, while the other is a moving-knife procedure that requires a referee to move one of the knives and that cannot be turned into an algorithm in the narrow sense of the word, but rather requires the participating agents to constantly monitor the moving knives and to react to their position in real time.

\paragraph{Selfridge-Conway procedure.}
The first envy-free procedures for 3 players has been proposed independently by Selfridge and Conway around~1960.

\begin{cakeproc}{Selfridge-Conway Procedure}{$3$}
\begin{enumerate}
\item Player~1 cuts the cake in three pieces (she considers equal).
\item Player~2 either ``passes'' (if she thinks at least two pieces
are tied for largest) or trims one piece (to get two tied for largest pieces).
--- \\ If she passed, then let players 3, 2, 1 pick (in that order). $\checkmark$
\item If player~2 did trim, then let 3, 2, 1 pick (in that order),
but require 2 to take the trimmed piece (unless 3 did). 
Keep the trimmings unallocated for now (note: the partial allocation is envy-free).
\item Now divide the trimmings.
Whoever of 2 and 3 received the \textit{un}trimmed piece does the cutting.
Let players choose in this order: non-cutter, player~1, cutter. $\checkmark$
\end{enumerate}\ \\[-35pt]
\end{cakeproc}
The Selfridge-Conway procedure is an algorithm in the proper sense of the word that always produces cake divisions that are envy-free, and thereby also proportional. It does not (always) produce contiguous pieces, but the number of cuts required is at most~5. There is no need for the active involvement of an external referee.

\paragraph{Stromquist procedure.}
The other classic procedure for envy-free cake-cutting has been proposed by Stromquist in~1980. It is a moving-knife procedure.

\begin{cakeproc}{Stromquist Procedure}{$3$}
\begin{enumerate}
\item A referee slowly moves a knife across the cake, from left to right
(supposed to cut somewhere around the $1/3$ mark). 
\item At the same time, each player is moving her own knife so that it would
cut the righthand piece in half (wrt.\ her own valuation).
\item The first player to call ``stop'' receives the piece to the left of
the referee's knife. The righthand part is cut by the middle one of the three
player knifes. If neither of the other two players hold the middle knife, they each obtain the piece at which their knife is pointing. If one of them does hold the middle knife, then the other one gets the piece at which her knife is pointing. $\checkmark$
\end{enumerate}\ \\[-35pt]
\end{cakeproc}
The procedure guarantees envy-free (and proportional) outcomes---which, it should be remarked, is not entirely obvious. To understand that it does indeed ensure envy-freeness, note that a player planning to shout ``stop'' at a certain point knows exactly who will receive which piece (so the players are not giving up control by claiming a particular piece of the cake). 

Being a moving-knife procedure, the Stromquist procedure does not qualify as an algorithm proper, and unlike for the Dubins-Spanier procedure there is no known ``discretisation''. The procedure clearly requires the active participation of a referee. On the upside, it does produce contiguous divisions.

\subsection{Overview of Procedures and Properties}

A summary of the properties of all the procedures covered is given by Table~\ref{tab:cakes}. Recall that envy-freeness entails proportionality. Also recall that the Banach-Knaster procedure does not produce contiguous divisions \textit{per se}, but a simple refinement of the procedure does. The two moving-knife procedures are the only procedures that require an active referee, although we have seen that for the Dubins-Spanier procedure there is a simple way of discretising the procedure that will remove the need for both the moving knife and the referee.

\begin{table}
\mbox{}\hfill\begin{tabular}{@{}|l|l|l|l|l|l|l|}\hline
\textbf{Procedure} & \textbf{Players} & \textbf{Type} & 
\textbf{Division} & \textbf{Contiguous?} & \textbf{Cuts} \\ \hline\hline
Cut-and-choose & $n=2$ & protocol & envy-free & yes & minimal \\ \hline
Steinhaus & $n=3$ & protocol & proportional & no & $\leq 3$ \\ \hline
Banach-Knaster & any $n$ & protocol & proportional & no/yes & bounded \\ \hline
Dubins-Spanier & any $n$ & 1 knife & proportional & yes & minimal \\ \hline
Even-Paz & any $n$ & protocol & proportional & yes & $O(n\log n)$ \\ \hline
Selfridge-Conway & $n=3$ & protocol &  envy-free & no & $\leq 5$ \\ \hline
Stromquist & $n=3$ & 4 knives & envy-free & yes & minimal \\ \hline
\end{tabular}\hfill\mbox{}
\caption{Cake-Cutting Procedures\label{tab:cakes}}
\end{table}

\subsection{Bibliographic Notes}

There are two excellent treatments of cake-cutting procedures available in book form, one by \citet{BramsTaylor1996} and the other by \citet{RobertsonWebb1998}, with the latter being somewhat more formal in style. References to the original papers proposing the various procedures discussed can be found in those books.
Another very helpful reference is the article by \citet{BramsTaylorAMM1995}, which besides introducing their (unbounded) algorithm for envy-free divisions amongst four (or more) players does an excellent job at presenting many of the classical procedures in a systematic manner. (The presentation of those procedures in the Notes you have in front of you owes much to that paper.) 

Studies of the complexity of cake-cutting include, besides the original contribution of \citet{EvenPaz1984}, work by \citet{WoegingerSgallDO2007}, \citet{EdmondsPruhsSODA2006}, and \citet{ProcacciaIJCAI2009}, as well as \citet{RobertsonWebb1998}. 

\subsection{Exercises}

\begin{exercise}
How many cuts are required, in the worst case, when $n$ players execute the
Banach-Knaster last-diminisher procedure to fairly divide a cake?
Justify your answer.
\end{exercise}

\begin{exercise}\label{ex:even-paz}
Show that $O(n\log n)$ individual marking queries are required when $n$ players execute the Even-Paz divide-and-conquer procedure. 
\end{exercise}

\begin{exercise}
Describe a discrete procedure for dividing a cake 
between four players that guarantees that each player believes 
they received at least $1/6$ of the cake and that uses only three cuts. 
(Additional marking queries as well as moving knives are not allowed.)
\emph{Adapted from \citet{RobertsonWebb1998}.}
\end{exercise}

\section{Indivisible Goods: Combinatorial Optimisation}
\label{sec:indivisible}

In this section, we shall discuss the problem of fair allocation when goods are \emph{indivisible} (and cannot be shared amongst several agents). Finding an allocation that is optimal with respect to a chosen fairness (or efficiency) criterion is then a combinatorial optimisation problem. We shall exemplify both centralised and distributed approaches to solving such a problem.

\subsection{The Model}
\label{sec:indivisible-model}

Let $\mathcal{N}=\{1,\ldots,n\}$ be a set of \emph{agents} and let $\mathcal{G}$ be a finite set of indivisible \emph{goods}. An \emph{allocation} $A:\mathcal{N}\to 2^{\mathcal{G}}$ is a mapping from agents to bundles of goods such that $A(i)\cap A(j)=\emptyset$ for all $i,j\in\mathcal{N}$ and $A(1)\cup\cdots\cup A(n)=\mathcal{G}$. That is, goods cannot be shared and we are interested in complete allocations. A state of the world is characterised by an allocation $A$ and a \emph{payment balance} $\pi:\mathcal{N}\to\R$, specifying for each agent an amount of money they are paying (or receiving, if $\pi(i)$ is negative), satisfying $\pi(1)+\cdots+\pi(n)=0$. That is, the overall balance of payments is always equal to~$0$.

Each agent $i\in\mathcal{N}$ has got a utility function $u_i:2^{\mathcal{G}}\times\R\to\R$ mapping states of the world to the reals. We assume that utility functions are \emph{quasi-linear}, i.e., $u_i(B,0)-u_i(B,x)=x$ for any bundle $B\in 2^{\mathcal{G}}$ and any payment $x\in\R$. We model utility functions by means of \emph{valuation functions} $v_i:2^{\mathcal{G}}\to\R$ mapping bundles of goods to the reals: $u_i(B,x)=v_i(B)-x$. 
We write $v_i(A)$ as a shorthand for $v_i(A(i))$.

For fair division of indivisible goods \emph{without money}, simply assume that payment balances are always equal to~$0$. In that case valuation and utility, as defined here, coincide.
 
\subsection{Preference Representation Languages}
\label{sec:preference-languages}

An important topic that we will \emph{not} cover in much detail in these Notes is the \emph{representation} of agent preferences. So far, we have only stipulated that each agent has got a utility function to model their preferences, but we have not said \emph{how} to encode this function in practice. 

\paragraph{The need for compact representation languages.}
To fully specify a fair division problem (and to allow for any kind of implementation) we also have to fix a concrete \emph{language} for describing valuation functions. The simplest approach is to use an \emph{explicit representation:} to represent a particular function over bundles of goods, use a big table to store for each bundle the value assigned to it.\footnote{Note that this is only possible if the number of bundles is finite. For the domain of cake-cutting discussed in Section~\ref{sec:cakes} there can be no general (finitary) language for describing all preferences that agents may have.} Of course, 
this will not be a very compact form of representation (in the case of, say, 20 indivisible goods, such a table can have over a million entries, because there are $2^{20}>1000000$ subsets of the full set of goods).

\paragraph{References for specific languages.}
Several languages for the \emph{compact} representation of preferences have been proposed in the literature: Languages for representing utility/valuation functions include \emph{weighted goals} 
\citep{UckelmanEtAlMLQ2009}, the \emph{$k$-additive form} \citep{ChevaleyreEtAlAOR2008}, \emph{bidding languages} developed in the combinatorial auction literature for modelling bids, such as the OR/XOR family of languages \citep{NisanCA2006}, and \emph{program-based representations} \citep{DunneEtAlAIJ2005}. 
Languages for representing ordinal preference relations include \emph{prioritised goals} \citep{LangAMAI2004}, \emph{CP-nets} \citep{BoutilierEtAlJAIR2004}, and \emph{CI-nets} \citep{BouveretEtAlIJCAI2009}.  

\paragraph{Survey papers.}
\citet{ChevaleyreEtAlAIMag2008} give an introduction to the problem of preference representation in the context of group decision making (which includes fair division), and the MARA Survey \citep{mara-survey} includes a review of preference representation languages that are relevant to multiagent resource allocation.

\subsection{Computational Complexity}
\label{sec:complexity}


One line of investigation that will be of interest to researchers in Computer Science and Multiagent Systems concerns the \emph{computational complexity} of fair division, and particularly so for the case of indivisible goods. 

\paragraph{Types of problems.}
For instance, we may ask how hard it is to check whether there exists an allocation with egalitarian social welfare above a certain threshold~$K$ (typically NP-complete, depending on the precise assumptions being made regarding the range of possible preferences and their representation), whether a given allocation is Pareto efficient (typically coNP-complete), or whether a given scenario admits a solution that is both envy-free and Pareto efficient (typically not even in NP). As many of these problems are computationally intractable, questions regarding the complexity of approximation schemes are also of interest. 

\paragraph{Problem representation.}
Importantly, any question concerning the complexity of a problem must be asked with respect to a particular form of encoding that problem. If the chosen encoding is not compact, then a result claiming low complexity (with respect to the size of the input) is not very informative or relevant. For fair division problems this means that we have to specify the language chosen for representing agent preferences when defining a problem the complexity of which we wish to study (see Section~\ref{sec:preference-languages}).

\paragraph{References.}
Important contributions to the literature studying the complexity of fair division with indivisible goods include those of  \citet{BouveretLangJAIR2008}, \citet{DunneEtAlAIJ2005}, and \citet{LiptonEtAlEC2004}. We will not survey this line of work in detail here; some of it has been reviewed in the MARA Survey \citep{mara-survey}.

\subsection{Centralised Algorithms}

Suppose we are given a set of agents, a set of goods, and the agents' utility functions over those goods (no money). We want to design an algorithm that uses this information to compute an allocation that is optimal in view of a given fairness criterion. 

\paragraph{Comparison with combinatorial auctions.}
From an algorithmic perspective, this problem is similar to the \emph{winner determination problem} in combinatorial auctions. In an auction, agents state their preferences by means of submitting bids to the auctioneer. In the standard model, the auctioneer is trying to find an allocation of goods to bidders that will maximise the sum of the prices specified via the bids. If we think of these prices as reflecting utility, then this amounts to finding an allocation that maximises utilitarian social welfare. This correspondence suggests that some of the algorithmic techniques employed (and developed) for combinatorial auctions can also be applied to the much wider range of fair division problems. These techniques include, in particular, integer programming (IP) and heuristic-guided search methods developed in~AI.
 
\paragraph{Maximising egalitarian social welfare.}
As an example, we give an algorithm for computing an allocation that maximises egalitarian social welfare using the framework of (mixed) IP. Before such an algorithm can be stated, we need to fix a language for representing the utility functions of the agents. To keep things simple, we shall be using an explicit form of representation, also known as the XOR-language in the combinatorial auctions literature. (We stress that this is not a very attractive language for most applications, because it is not a compact representation language, but it does have the advantage of being very simple.) Each agent provides a list of bundles they are interested in, each labelled with a utility value. We can award each agent at most one of the bundles they listed, and they will derive from it the utility specified. 

Let $\langle B_{ij},u_{ij}\rangle$ the $j$th labelled bundle representing the utility function of agent~$i$. Let $n_i$ be the number of bundles listed by agent~$i$. Introduce a binary decision variable $x_{ij}$ for every $i\in\mathcal{N}$ and $j\leq n_i$, with $x_{ij}=1$ if agent~$i$ does receive bundle $B_{ij}$ and $x_{ij}=0$ otherwise. Let $m$ be the number of goods. 
We need to satisfy the following constraints.
\begin{itemize}
\item First, ensure that every item gets allocated to at most one agent (that is, we do allow for \emph{free disposal}). In the next set of constraints, the constant $[k\in B_{ij}]$ is $1$ if the $k$th item is contained in bundle $B_{ij}$, and $0$ otherwise. 
\begin{eqnarray}\label{eq:allocation-constraint}
(\forall k\leq m)\quad \sum_{i\in\mathcal{N}}\sum_{j=1}^{n_i} [k\in B_{ij}]\cdot x_{ij}& \leq & 1
\end{eqnarray}
\item Second, ensure that each agent receives at most one of the bundles they specified.
\begin{eqnarray}
(\forall i\in\mathcal{N})\quad \sum_{j=1}^{n_i} x_{ij}& \leq & 1
\end{eqnarray}
Note that we can now write the utility enjoyed by agent~$i$ as $\sum_{j=1}^{n_i}u_{ij}\cdot x_{ij}$. Above constraint guarantees that at most one of the summands will be different from zero. 
\item We now introduce one further decision variable, $y$ (which need not take integer values, so we are operating in the realms of \emph{mixed} IP). It is used to represent an upper bound on the egalitarian social welfare of the allocation represented by the instantiations of the variables of type $x_{ij}$. 
\begin{eqnarray}\label{eq:bound-constraint}
(\forall i\in\mathcal{N})\quad y & \leq & \sum_{j=1}^{n_i}u_{ij}\cdot x_{ij}
\end{eqnarray}
\end{itemize}
Finding an allocation with maximal egalitarian social welfare now amounts to solving the following mixed IP:
\begin{equation}
\mbox{maximise}\  y \quad\mbox{subject to constraints (\ref{eq:allocation-constraint})--(\ref{eq:bound-constraint})}
\end{equation} 
The optimal allocation can be read off the instantiation of the variables $x_{ij}$: item~$k$ should be given to agent~$i$ if and only if there exists a $j$ such that $x_{ij}=1$ and item~$k$ is contained in the set~$B_{ij}$.

Similar algorithms can  be designed for (some of) the other fairness and efficiency criteria reviewed in Section~\ref{sec:criteria}. In particular, for the utilitarian CUF this has been done in the combinatorial auction literature for a range of different preference representation languages. 

\subsection{Distributed Negotiation}

The centralised approach is attractive in that it gives maximum flexibility to the algorithm designer. But it also has a number of disadvantages. First and foremost, it presupposes the availability of a centre that can carry out the (often costly) computations required to determine a socially optimal allocation. Furthermore, the agents involved must be able to trust that centre. Arguably, the problem of finding a fair allocation presents itself much more naturally as a distributed problem, because the information regarding individual preferences is distributed amongst the agents. 

In a \emph{distributed approach} to fair division, (small) groups of agents can locally agree on a sequence of deals to exchange some of goods held by them by means of negotiation. There is no global control determining which agents have to interact in which manner at any given time.\footnote{The interactive procedures for fairly dividing a cake discussed in Section~\ref{sec:cakes} fall somewhere between the centralised and the distributed approach as presented here (although they are, arguably, closer to the centralised approach). They are not distributed, because they all come with a fixed protocol that determines which agent will have to answer which query at any given time. On the other hand, they are also not entirely centralised, because it is not the case that agents first communicate their complete preferences and the centre then computes a fair allocation, but agents only answer very specific queries regarding their preferences, from which the final division follows almost immediately.}
The question then arises whether we can endow individual agents with a suitable set of rules determining their willingness to accept certain deals such that resulting allocations will satisfy our fairness or efficiency criterion of choice (\emph{``welfare engineering''}). Or, \textit{vice versa}, given certain assumptions on the local behaviour of agents, we may ask to what extent we can predict the properties of the allocations emerging at the global level. 

We shall go through one example that shows that under reasonable assumptions on the behaviour of individual agents, we can set up a distributed protocol that will ensure that a socially desirable allocation, in this case an allocation maximising utilitarian social welfare, will emerge eventually. (The result may also be interpreted the other way round: it shows how to design your agents so as to guarantee that, when they are left alone to negotiate autonomously, an optimal allocation will be reached eventually.)

\paragraph{Deals and myopic individual rationality.}
Given an initial allocation of goods to agents, agents negotiate a sequence of deals. A deal $\delta=(A,A')$ is a pair of allocations, describing the situation before and after the deal. We do not put any structural restrictions on deals: any number of items may be exchanged between any number of agents in a single deal. Each deal will be paired with a set of monetary side payments, represented by a \emph{payment function} $p:\mathcal{N}\to\R$ satisfying $p(1)+\cdots+p(n)=0$. (The payment balance $\pi$ is the sum of all the payment functions associated with the deals made in the past.)

If agents are selfish, a reasonable assumption is that any deal they are prepared to accept will improve their own utility by at least a small margin. They will never accept a deal that leaves them worse off. This assumption will not be satisfied under \emph{all} circumstances, as an agent may be prepared to accept a temporary loss in utility, if this promises to open up opportunities for future gains, but for agents that are either very risk-averse or that are not capable of reasoning about the future, this kind of myopic interpretation of rationality is reasonable. We now define the class of deals that such agents would be prepared to accept.

\begin{definition}[Individual rationality]
A deal $\delta=(A,A')$ is myopically individually rational (IR) if there exists a payment function $p$ such that $v_i(A')-v_i(A)>p(i)$ for all agents $i\in\mathcal{N}$, except possibly $p(i)=0$ for agents~$i$ with $A(i)=A'(i)$.
\end{definition}
That is, for each agent, her gain in valuation must outweigh her payment (or, if she is losing valuation, her loss in valuation must be outweighed by the payment received by her). Only agents not involved in the deal (any agent whose allocated bundle does not change) may receive a payment of~$0$.

\paragraph{Convergence.}
While individual rationality can be (and has been) motivated purely from the local viewpoint of individual agents as a reasonable criterion for selfish but myopic agents to decide which deals to accept, it turns out to also be very helpful in negotiating socially efficient allocations. In fact, as we shall see next, if myopic IR agents are left to their own devices, then \emph{any sequence of deals that they may agree on will converge to an allocation with maximal utilitarian social welfare}. This surprising result can largely be explained by means of the following lemma \citep[Lemma~1]{EndrissEtAlJAIR2006}.

\begin{lemma}
\label{lem:irsw}
A deal $\delta=(A,A')$ is IR if and only if $\sw{util}(A)<\sw{util}(A')$.
\end{lemma}
That is, a deal is IR if and only if it increases utilitarian social welfare.
Convergence then follows almost immediately: 
The number of possible allocations is finite, because both $\mathcal{N}$ and $\mathcal{G}$ are.
By Lemma~\ref{lem:irsw}, any IR deal must strictly increase social welfare. 
Hence, we can never visit a single allocation twice, i.e., negotiation must eventually terminate.
The terminal allocation must have maximal utilitarian social welfare, because if it did not, then by Lemma~\ref{lem:irsw}, any deal taking us to an allocation with greater social welfare would be IR and thus available to the agents (thereby contradicting the assumption that termination has already occurred). 

Convergence means that we do not need to coordinate the behaviour of individual agents. Provided they can be expected to identify new mutually beneficial deals as long as such deals are possible at all, we have a formal guarantee that an optimal agreement will eventually be found. However, the problem of finding such an optimal allocation is still NP-hard, which manifests itself in the fact that the sequence of deals required may be very long and some of the individual deals may need to have very high structural complexity (involving many agents and/or goods). Indeed, there are results that show that deals involving any number of agents and any number of goods will be necessary in some cases (for some combinations of valuation functions and initial allocation). On the other hand, under additional assumptions on the valuation functions (in particular when they are all additive), structurally simpler deals will sometimes be sufficient to guarantee convergence.

A similar programme can (and, to some extent, has been) carried out for other fairness and efficiency criteria. For any given criterion, we can seek to design a local rationality criterion for agents to decide which deals to accept, given the information locally available to them. In some cases it will then be possible to establish convergence results similar to the above, which guarantee socially optimal outcomes by means of autonomous negotiation amongst the agents thus designed.

\subsection{Bibliographic Notes}

In the section on centralised algorithms, we have drawn some parallels to combinatorial auctions. The combinatorial auction handbook \citep{CramtonEtAlCA2006} presents combinatorial auctions and algorithms for solving them in detail. \citet{BouveretLemaitreAIJ2009} give (centralised) algorithms for computing a leximin-optimal solution. 

A starting point for finding out more about the distributed approach to multiagent resource allocation is \citep{EndrissEtAlJAIR2006}. The result on convergence to an allocation with maximal utilitarian social welfare by means of individually rational deal is originally due to \citet{SandholmAAAISS1998}.
Other work has discussed the \emph{communication complexity} of negotiation, and established bounds on the number of local deals required to reach a socially optimal allocation \citep{EndrissMaudetJAAMAS2005,DunneJAIR2005}.

\subsection{Exercises}

\begin{exercise}
What is the computational complexity of (the decision variant of) the problem of finding an allocation of indivisible goods (without money) to agents that maximises elitist social welfare?
\begin{enumerate}
\item First state your answer (and your proof) with respect to the explicit form of representing utility functions (where the size of the representation of a function is proportional to the number of bundles to which it assigns a non-zero value).
\item Then repeat the same exercise, this time assuming that utility functions are expressed using the language of weighted propositional formulas \citep{UckelmanEtAlMLQ2009}.
\end{enumerate}
\end{exercise}

\begin{exercise}
Devise an IP algorithm for computing an allocation with maximal Nash social welfare when utility functions are represented using the explicit form. 
\end{exercise}

\section{Conclusion} 

The goal of these notes has been to provide some basic background (and references for further reading) on fair division to people interested in using these concepts for their research in Artificial Intelligence, Multiagent Systems, Computational Social Choice, or similar, as well as to point researchers in these areas to topics in fair division where the application of tools and techniques from Computer Science may still yield new and interesting results.

Section~\ref{sec:criteria} can serve as a catalogue of criteria regarding the economic quality of an allocation of resources that someone developing a multiagent system may wish to consult before specifying the precise goals of their system. As mentioned in the text, new applications may very well call for new such criteria, but these should always be formulated in relation and with reference to the standard criteria for which strong philosophical justifications and deep mathematical results are available.

Sections~\ref{sec:cakes} and~\ref{sec:indivisible} showcase some examples for algorithmic approaches of varying sorts to the problem of fairly dividing a number of goods. There still is a lot of room for improvement, and hopefully these examples can help inspire new ideas.


\end{document}